\pdfoutput=1

\documentclass[11pt,table]{article}

\usepackage[]{EMNLP2023}

\usepackage{times}
\usepackage{soul}
\usepackage{latexsym}
\usepackage{tabularx}
\usepackage{hyperref}
\usepackage{url}
\usepackage{graphicx}
\usepackage{amsmath}
\usepackage{siunitx}
\usepackage{booktabs}
\usepackage{multirow}
\usepackage{arydshln}
\usepackage{enumitem}
\usepackage{xargs}
\usepackage{svg}
\usepackage{stfloats}
\usepackage{marginnote}
\usepackage{amssymb}
\usepackage{appendix}
\usepackage{array}
\usepackage{fancyvrb}
\usepackage{caption}
\usepackage{subcaption}
\usepackage{pifont}
\usepackage{makecell}
\usepackage[section]{placeins}
\usepackage{framed,verbatim}
\usepackage{alltt}

\usepackage[T1]{fontenc}

\usepackage[utf8]{inputenc}

\usepackage{microtype}
\usepackage{inconsolata}

\title{DUBLIN: Visual Document Understanding By Language-Image Network }

\author{{\bf Kriti Aggarwal\thanks{\enspace Equal contribution}, }{\bf Aditi Khandelwal\footnotemark[1]}, {\bf Kumar Tanmay\footnotemark[1]},\\ {\bf Owais Mohammed Khan}, {\bf  Qiang Liu}, {\bf Monojit Choudhury}\textbf{,}\\ {\bf Hardik Hansrajbhai Chauhan}\textbf{,} {\bf Subhojit Som}\textbf{,}{\bf  Vishrav Chaudhary}\textbf{,} {\bf Saurabh Tiwary} \\
        Microsoft Corporation\\
        {\{kragga, t-aditikh, t-ktanmay, owais.mohammed, qiangliu\}}@microsoft.com, \\
        \{monojitc, hachauhan, subhojit.som, vchaudhary, satiwary\}@microsoft.com}

\begin{document}
\maketitle
\begin{abstract}

In this paper, we present DUBLIN, a pixel-based model for visual document understanding that does not rely on OCR. DUBLIN can process both images and texts in documents just by the pixels and handle diverse document types and tasks. DUBLIN is pretrained on a large corpus of document images with novel tasks that enhance its visual and linguistic abilities. We evaluate DUBLIN on various benchmarks and show that it achieves state-of-the-art performance on extractive tasks such as DocVQA, InfoVQA, AI2D, OCR-VQA, RefExp, and CORD, as well as strong performance on abstraction datasets such as VisualMRC and text captioning. Our model demonstrates the potential of OCR-free document processing and opens new avenues for applications and research.

\end{abstract}

\section{Introduction}
\label{sec:introduction}

Humans have an incredible ability to process documents visually, interpreting the layout and extracting valuable information from images and texts simultaneously. Document layouts, with strategically placed figures, tables, and other visual elements, are designed to cater to human perception and visual cognition biases \cite{kress2020reading}.  However, in most contemporary visual document processing models, on the other hand, OCR is commonly employed as a starting point \cite{LayoutLM,xu2022layoutlmv2,huang2022layoutlmv3,peng2022ernielayout} for extracting the text, followed by a text-only processing scheme. Despite its usefulness, OCR can introduce errors, which can be particularly problematic in scenarios involving non-Latin scripts or handwritten content. More importantly, OCR-based methods fall short in capturing the rich visual context present in document images, making them less effective for various applications \cite{ocrerror1,ocrerror2,ocrerror3}.

Previous attempts to address OCR-related limitations have led to the emergence of models such as Donut~\cite{donut} and Pix2struct~\cite{lee2022pix2struct}, which aim to process documents without relying on OCR. Although these models hold promise, their applications have been somewhat limited, and they do not fully exploit the potential of visual document understanding. While Donut performs well on data resembling their pretraining samples, it shows poor performance when tested on datasets with complex documents such as InfoVQA. Pix2struct lacks thorough evaluation on diverse tasks such as information extraction, table question answering, and machine reading comprehension (MRC), leaving questions about its versatility unanswered.

To overcome the aforementioned challenges and advance the field of visual document understanding, we present \textbf{DUBLIN}:  Visual \textbf{D}ocument \textbf{U}nderstanding \textbf{B}y \textbf{L}anguage-\textbf{I}mage \textbf{N}etwork, a generic pixel-based approach to achieve OCR-free document processing without the need for any specialized pipelines. DUBLIN achieves state-of-the-art performance on extractive tasks, including Document-based visual question-answering (DocVQA - 5.35\% $\uparrow$), (InfographicsQA - 7.5\% $\uparrow$), QA over illustrations (AI2D - 24\% $\uparrow$, OCR-VQA - 3.8\% $\uparrow$), UI understanding (RefExp - 5\% $\uparrow$), and information extraction (CORD - 6\% $\uparrow$). Additionally, it demonstrates strong performance on abstraction tasks such as machine reading comprehension (VisualMRC - $1\%$$\uparrow$) and text captioning of natural images. Furthermore, our model achieves competitive performance with existing approaches on tasks like table question-answering, document classification, and web-based structured reading comprehension.

Our model showcases adaptability and versatility, which are attributed to a carefully designed pretraining recipe. By employing curriculum learning and incorporating novel tasks like bounding box task, rendered question-answering task, and masked document language modeling task during pretraining, our model acquires the ability to seamlessly integrate new tasks and achieve state-of-the-art (SOTA) performance across various document understanding tasks. Our contributions extend the possibilities for applications, from search engines to presentations, and we hope our work will inspire further developments in the field of visual document processing.

\section{Related Works}
\label{sec:relatedworks}

The transformer architecture has become prevalent in document understanding, and the LayoutLM family of models has extended transformer-based approaches like BERT \cite{devlin2019bert} to handle document visuals. Various features, such as 2D spatial positional information \cite{LayoutLM}, visual tokens, spatially biased attention \cite{xu2022layoutlmv2}, and crossmodal alignment objective \cite{huang2022layoutlmv3}, have been integrated into these models. However, some evaluations of LayoutLM models overlooked text recognition, an essential task.
DocFormer used only visual features near text tokens \cite{appalaraju2021docformer}. Ernie-Layout used reading order prediction as a pretraining task \cite{peng2022ernielayout}. TILT trained generative language models on document data using generative objectives \cite{TILT}.

Recent advances in document understanding have focused on self-supervised learning and multi-modal embeddings. UDoc used multi-modal embeddings and self-supervised losses to learn joint representations for words and visual features from document images \cite{udoc}. SelfDoc used coarse-grained multimodal inputs, cross-modal learning, and modality-adaptive attention to model document components \cite{li2021selfdoc}. UDOP used a Vision-Text-Layout Transformer and a prompt-based sequence generation scheme to enable document understanding, generation, and editing across domains \cite{udop}.

The above-described models depend on off-the-shelf OCR tools for text processing in documents, which limits their applications and increases computational costs. Recent models like Donut \cite{donut}, Dessurt \cite{dessurt}, and Pix2Struct \cite{lee2022pix2struct} are end-to-end image-to-text models that do not need OCR at inference time. Pix2struct is a pretrained image-to-text model for purely visual language understanding that can be fine-tuned on tasks containing visually-situated language \cite{lee2022pix2struct}. It was pretrained by learning to parse masked screenshots of web pages into simplified HTML and enables resolution flexibility to a variety of visual language domains. Matcha proposed pretraining objectives to enhance the mathematical reasoning and chart derendering capability of visual language models \cite{liu2022matcha}.

\section{Method}
\label{sec:method}

\subsection{Model Architecture}

\begin{figure*}[t]
\includegraphics[width=\textwidth]{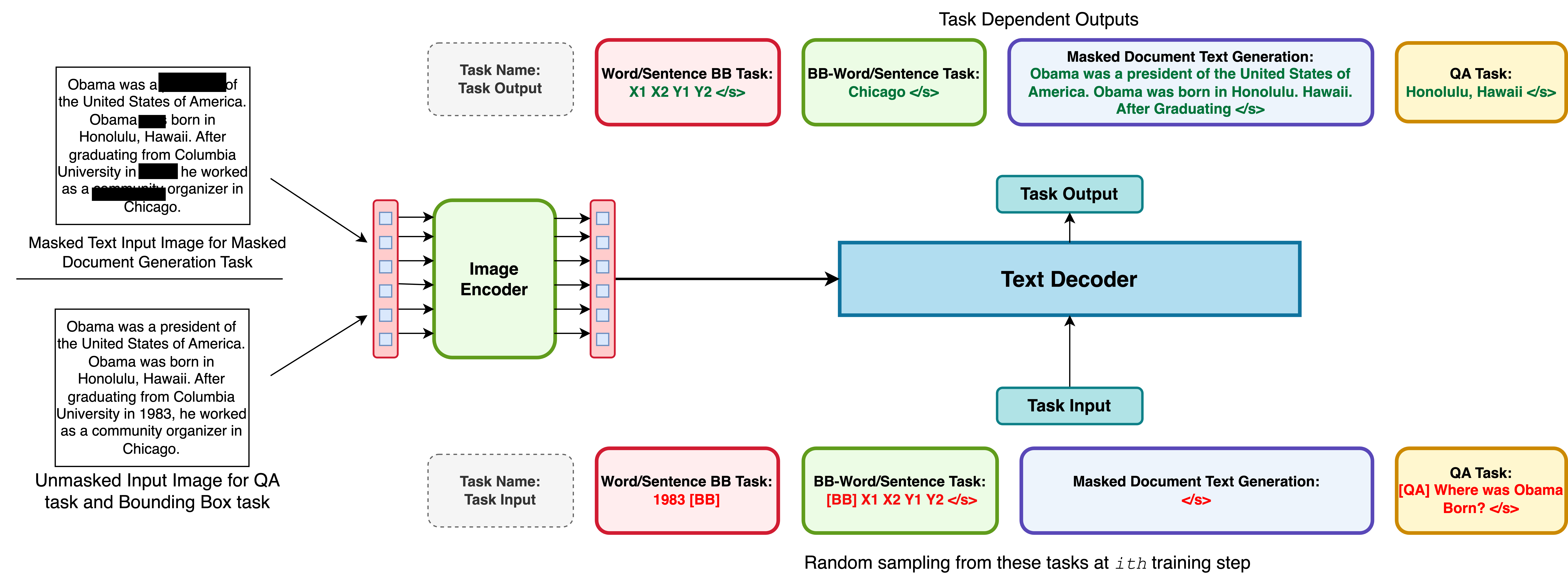}
\caption{Illustration of three tasks in the DUBLIN pretraining framework: Bounding Box, Rendered QA, and Masked Document Text Generation.}
\label{fig:pretraining_tasks}
\end{figure*}

DUBLIN is a novel end-to-end framework that combines the Bletchley \cite{Bletchley} image encoder and the text decoder initialized by the weights from InfoXLM's text encoder \cite{chi2021infoxlm}. Bletchley is a multimodal model that employs a bootstrapping mechanism to train image and text encoders that can handle different modalities. InfoXLM is a cross-lingual model that learns a universal language representation that can handle diverse languages. Our model has 976M trainable parameters and incorporates cross-attention layers between the image encoder and the text decoder to model the interaction between the visual and textual modalities. This enables the decoder to attend to pertinent regions in the image based on the query or context. We adopt Bletchley’s image encoder and InfoXLM’s text encoder as the initial weights for our model and then further pretrain them on various datasets using a combination of multi-task pretraining objectives and curriculum learning. The pretraining datasets comprise CCNews 200M \cite{wenzek-etal-2020-ccnet}, Google NQ Dataset \cite{googleNQ}, Microsoft Bing QA Dataset, Rendered InfoXLM EN Dataset \cite{chi2021infoxlm}, and Synthetic Table QA, which are detailed further in Section~\ref{sec:pretraindata}.

\subsection{Pretraining Objectives}
We propose a novel pretraining framework with four objectives at different levels: language, image, document structure, and question-answering. These objectives aim to capture the complex structures of visual documents and enhance the model’s holistic comprehension and reasoning abilities. Figure~\ref{fig:pretraining_tasks} shows the generative pretraining tasks for DUBLIN. We describe the pretraining objectives below.

\noindent \textbf{Masked Document Language Modeling Task}~We propose a pretraining objective that leverages both image and text modalities to learn a cross-modal representation for document understanding. Our objective consists of masking 15\% of the text regions randomly in the document image and masking the corresponding text tokens in the sequence formed by concatenating all the text in the image. The text decoder then tries to predict the masked text tokens, given the masked document image and the unmasked text tokens as contexts. The image encoder encodes the masked image into a sequence of hidden states, which are used by the cross-attention mechanism in the text decoder to align the image and text modalities. We use the cross-entropy loss as our loss function to measure the difference between the predicted and true text tokens. By doing so, our model learns to read and understand the text from the document image, as well as capture the cross-modal dependencies.

\noindent \textbf{Bounding Box Task.}~We also propose a bounding box task to learn the location and content of text regions in the document image. For this task, we encode the text and the top left and bottom right coordinates of its bounding box using a special token format. For instance, the sequence \texttt{<s>} \texttt{text} \texttt{</s>} \texttt{[BB] x1  y1  x2  y2} is used to predict the text’s bounding box, while the sequence \texttt{[BB] x1  y1  x2  y2} \texttt{<s>} \texttt{text} \texttt{</s>}  is used to predict the text within the bounding box. We adopt cross-entropy loss as our loss function for this task. This task enables our model to localize and recognize the text regions in the document image.

\noindent \textbf{Rendered Question Answering Task.}~We introduce this task specifically to aid the model in document image question answering. Using publicly available text QA datasets -- Rendered InfoXLM EN Dataset \cite{chi2021infoxlm}, and Google NQ \cite{googleNQ}, as well as two proprietary datasets based on Web QA and synthetically generated table QA (the datasets are described in the next section) we created instances of visual QA task by rendering the passage and question as an image and input it to the image encoder. We use the question as the prefix for the text decoder to generate the answer. We use the cross-entropy loss function for this task.

\noindent \textbf{Masked Autoencoding Task.}~
Following ViT-MAE \cite{he2021masked}, we use the MAE task as the initial pre-training objective to train the image encoder prior to the above three strategies. This is done by reconstructing 15\% randomly masked image patches with the help of an equivalent image decoder. We use 1-D fixed sinusoidal position embeddings and a normalized MSE pixel reconstruction loss for this task. Additional details can be found in Appendix~\ref{A:pretraining_tasks}.

\subsection{Pretraining Data}
\label{sec:pretraindata}

To pretrain our model on various tasks, we use five datasets: CCNews 200M \cite{wenzek-etal-2020-ccnet}, Google NQ Dataset \cite{googleNQ}, Rendered InfoXLM EN Dataset \cite{chi2021infoxlm}, Bing QA Dataset, and Synthetic Table Structure QA Dataset. These datasets contain both text and image information, which we leverage to train our model on multimodal understanding and generation. For the CCNews 200M and Google NQ datasets, we use the Selenium tool to capture screenshots and texts along with their bounding boxes from the HTML documents. For the InfoXLM EN dataset, we render the text documents as images with different data augmentations such as random font, style and color. For the proprietary CSE QA dataset, we render the text document and a question together as an image. For the Synthetic Table Structure QA dataset, we generate synthetic questions and answers for the table structure task using templates. We provide more details about each dataset and the data processing steps in Appendix~\ref{A:pretrainingData}.

\noindent \textbf{Model Pretraining.} We use the XLM-RoBERTa tokenizer from the HuggingFace Transformers library and augment our vocabulary with special tokens: \texttt{<BB>}, \texttt{<QA>} and 1024 patch tokens. We use AdamW Optimizer with a learning rate of $1e^{-4}$, 10000 warmup steps, effective batch size of 1024 with low-resolution images and 256 with high-resolution images, weight decay of $0.01$, $\beta_1 = 0.9$ and $\beta_2 = 0.999$. The pretraining procedure consists of five stages, with each stage, adding new tasks/complexity to the training process.  In the first stage, we resize the input image to $224\times224$ and split it into fixed patches of $14\times14$ to feed to the image encoder. The model is then trained using MAE and Masked Document Language Modeling tasks simultaneously on low-resolution images sampled from the CCNews 200M, Google NQ, and Rendered InfoXLM EN datasets for 50k steps. In the second stage, we introduce the Rendered Question Answering Task using the Google NQ and Bing QA datasets for 350k steps at the same resolution. The third stage involves increasing the resolution to $896\times896$ and repeating the above two stages combined for 55k steps. The data will be sampled equally from each of the above four datasets. In the fourth stage, we add the bounding box prediction objectives and continue training for another 150k steps on high-resolution images (896×896). Finally, in the last stage of the curriculum, we include the Synthetic Table QA dataset and further pretrain our model for a total of 600k steps. Now we can use this pre-trained model to be finetuned on different downstream tasks.

\section{Experiments and Results}
\label{sec:finetuning}
We conduct comprehensive experiments on various types of documents, such as handwritten, typewritten, scanned, infographics, diagrams, tables, and webpages, and evaluate our model on various downstream tasks to assess the model's generalization capability. In this section, we describe the tasks, datasets, and the results. For each experiment, we finetune our pretrained model on a dataset and then report the performance. For each dataset, we use the publicly available train/development/test set splits, except for WebSRC where the test set is not released and hence we report performance on the development set. The hyperparameters used for finetuning are listed in Appendix \ref{a:hyperparams}.   We also adopt the following two generic strategies for input formatting:

First, inspired by the Pix2Struct, for all tasks, we append the question/key visually rendered onto the document image itself as can be seen in Figure~\ref{fig:questionrendered}. Subsequently, we also utilize the question/key as a prefix for the text decoder. 

Second, for accommodating diverse image size and aspect ratios, we employ a {\em Variable Resolution Finetuning} strategy. \citet{lee2022pix2struct} addresses the issue of aspect ratio distortion by rescaling input images either up or down to ensure the extraction of the maximum number of patches that can fit within the designated sequence length. However, this resizing technique can lead to a potential loss of information due to under-utilization of maximum sequence length tokens. In contrast, we focus on preserving information by adopting a different strategy that resizes the image to an aspect ratio which is an even power of 2 (e.g., 1, 4, 16, 64, etc.) as depicted in Figure~\ref{fig:varres}. By doing so, we maintain the desired aspect ratio while accommodating the maximum allowable number of patches (4096) within the given sequence length. 
As a result, we have two versions of DUBLIN, one fixed resolution model and one variable resolution model, which we called DUBLIN\textsubscript{fixed\_res} and DUBLIN\textsubscript{variable\_res} respectively.

\begin{table*}[t]
\centering
\small
\begin{tabular}{l  cccc c  cc cc}
\toprule
\multirow{1}{*}{Model} & \multicolumn{3}{c}{QA over Illustrations} &  \multicolumn{3}{c}{UI understanding} & \multicolumn{1}{c}{Captioning}  & \multicolumn{2}{c}{Document QA}\\
\cmidrule(lr){2-4} \cmidrule(lr){5-7} \cmidrule(lr){8-8} \cmidrule(lr){9-10}
 &  \textbf{ChQA} & \textbf{AI2D} & \textbf{O-VQA} & \textbf{RefExp} & \begin{tabular}[c]{@{}c@{}} \textbf{Widget} \\ \textbf{Cap} \end{tabular}  & \textbf{Scrn2Wds} & \textbf{TCaps} & \textbf{DVQA} & \textbf{IVQA} \\
\midrule
Metrics & RA & ANLS & F1 & EM & CIDEr & CIDEr & CIDEr & ANLS & ANLS \\
\midrule
Donut&  41.8 & 30.8 & 66.0 & - & 127.4 & 56.4 & 74.4  & 67.5 & 11.6\\
Pix2Struct\textsubscript{large}  & \textbf{58.6}&  42.1 & 71.3 & 94.2 & \textbf{136.7}  & \textbf{109.4} & \textbf{95.5} & 76.6 & 40.0\\
\rowcolor{cyan!10} Dublin\textsubscript{fixed\_res} &{35.6}  & \textbf{51.1} & \textbf{73.1} & \textbf{99.1}  & 132.2 & 101.8 & 92.8 & \textbf{78.2} & \textbf{36.8} \\
\rowcolor{cyan!10} \textbf{Dublin\textsubscript{variable\_res}} & {35.2} & \textbf{52.3} & \textbf{74.0} & \textbf{99.1} & 132.2 & 101.8 & 92.8  & \textbf{80.7} & \textbf{43.0}\\ 
\midrule{}
\begin{tabular}[c]{@{}c@{}}(SOTA with)\\ spl. pipelines\end{tabular} & \begin{tabular}[c]{@{}c@{}}(VTP)\\ 45.5\end{tabular} & \begin{tabular}[c]{@{}c@{}}(DQAN)\\ 38.5\end{tabular} & \begin{tabular}[c]{@{}c@{}}(LATr)\\ 67.5\end{tabular} & \begin{tabular}[c]{@{}c@{}}(UIB)\\ 90.8\end{tabular} & \begin{tabular}[c]{@{}c@{}}(VUT)\\ 97.0\end{tabular} & \begin{tabular}[c]{@{}c@{}}(VUT)\\ 64.3\end{tabular} & \begin{tabular}[c]{@{}c@{}}(PaLI)\\ 160.4 \end{tabular} & \begin{tabular}[c]{@{}c@{}}(UDOP)\\ 84.7\end{tabular} & \begin{tabular}[c]{@{}c@{}}(UDOP)\\ 47.4\end{tabular} \\
\bottomrule
\end{tabular}
\caption{
Performance on QA over illustrations, UI understanding, image captioning and QA tasks. Higher the better. ChQA: ChartQA, O-VQA: OCR-VQA, Scrn2Wds: Screen2Words, TCaps: Text Captioning, DVQA: DocVQA, IVQA: InfoVQA, VTP: Vision Tapas Model \cite{masry2022chartqa}, DQAN: Diagram Question-Answering Network \cite{kembhavi2016diagram}, LATr: Layout-Aware Transformer for Scene-Text VQA \cite{biten2021latr}, UIB: UI-Bert \cite{bai2021uibert}, VUT: Versatile UI Transformer \cite{li2021vut}, PaLI: Pathways Language and Image model  \cite{chen2023pali}.
}
\label{tab:resultsonvisualtasks}
\end{table*}

\subsection{Downstream Tasks}

\noindent \textbf{Question Answering.}~We utilize DocVQA \cite{mathew2021docvqa} and InfographicsVQA \cite{mathew2021infographicvqa} from the DUE benchmark \cite{duebenchmark} for document question-answering task. These datasets allow us to assess the performance of our model in question answering on documents and infographics, respectively. We evaluate our model's performance on the WebSRC dataset \cite{websrc} for webpage-based structural reading comprehension. For QA tasks related to illustrations, we test our model on ChartQA \cite{masry2022chartqa}, AI2D \cite{katti2018chargrid}, and OCR-VQA datasets \cite{ocr-vqa}. Additionally, we test DUBLIN's performance on Squad1.1 \cite{rajpurkar2016squad} by rendering the textual passage as images. More details about the datasets and preprocessing can be found in Appendix~\ref{A:finetuningdatasets}. 

\noindent \textbf{Information Extraction (IE).}~We leverage the  DeepForm dataset \cite{Deepform} from the Due Benchmark for the key information extraction task. To accomplish this task, we overlay the extracted key information on top of the corresponding image and utilize it as a prefix for the text decoder. We also test DUBLIN on two Information Extraction benchmarks: CORD  \cite{Park2019CORDAC} and FUNSD \cite{jaume2019funsd}. FUNSD is a BIO-scheme-based word-labeling task where the labels are semantic entity types: question, answer, header, or other. CORD is also a word-labeling task with 30 labels (fields) under 4 categories, which are key information from receipts. 

\noindent \textbf{Table Question Answering/NLI.} We utilize the WikiTable Questions dataset \cite{pasupat2015compositional} from the DUE benchmark and the WikiSQL QA dataset \cite{zhong2017seq2sql} for table-based QA. The WikiSQL dataset has tables in JSON format that we rendered as images in various styles. Additionally, we also test our model on the Tabfact dataset \cite{chen2020tabfact}, which requires a comprehensive understanding of the table content.

\noindent \textbf{Document Classification.}~To evaluate our model's performance on document classification, we conduct experiments on the RVL-CDIP dataset \cite{harley2015evaluation}. This dataset contains scanned document images categorized into 16 classes, including letters, forms, emails, resumes, memos, etc.

\noindent \textbf{UI Understanding}~For the UI understanding task, we evaluate on three datasets: RefExp \cite{bai2021uibert}, Widget Captioning \cite{li2020widget} and Screen2words \cite{wang2021screen2words}. In RefExp, the goal is to identify a specific component in an app using a natural language expression and a screenshot with highlighted bounding boxes. Widget Captioning involves describing a widget's functionality with a single bounding box, while Screen2Words focuses on captioning an entire page's functionality based on an app screenshot.

\noindent \textbf{Image Captioning}~We also show that our model can generate image captions by evaluating it on the TextCaps dataset \cite{sidorov2020textcaps}.

\noindent \textbf{Machine Reading Comprehension (MRC)}~ We utilize VisualMRC dataset \cite{tanaka2021visualmrc}, a webpage-based dataset where the model needs to give an abstractive answer based on the question for testing reading comprehension from images.

\subsection{Results}
\label{sec:results}
Since we have multiple tasks, we present the results in three task-wise tables: Table~\ref{tab:resultsonvisualtasks}, Table~\ref{tab:mixdatasets}, and Table~\ref{tab:TableQA}. Table~\ref{tab:resultsonvisualtasks} displays the results on QA over illustrations, UI understanding, Image Captioning, and Document QA tasks. In Table~\ref{tab:mixdatasets}, we showcase the results on IE, classification, and extractive and abstractive reading comprehension tasks. Table~\ref{tab:TableQA} contains the results for Table QA and rendered datasets. Tables~\ref{tab:resultsonvisualtasks} and \ref{tab:mixdatasets} show a comparison with pixel-based models in the first segment, and in the second segment, we report the current SOTA models with specialized pipelines and text-based baselines, if any. In Table~\ref{tab:TableQA}, we present DUBLIN's result in the first segment as there are no other pixel-based baselines and in the second segment we report the current SOTA models' performance and text-based baseline.

\begin{table*}[t]
\centering
\small
\begin{tabular}{lcccccc}
\toprule
\multirow{2}{*}{\textbf{Model}}                            & \multicolumn{3}{c}{\textbf{Information Extraction}}    & \textbf{Classification}                                & \multicolumn{2}{c}{\textbf{Reading Comprehension}}                                                                                     \\
\cmidrule(lr){2-4} \cmidrule(lr){5-5} \cmidrule(lr){6-7} 

& \textbf{FUNSD}    & \textbf{CORD} & \textbf{DeepForm} & \textbf{RVL-CDIP}  & \textbf{WebSRC} & \textbf{VisualMRC}                                              \\
\midrule
Metrics   & F1 & F1     & F1     & Accuracy  & EM/F1  & CIDEr\\
\midrule
Donut   & - & 91.6 & - & \textbf{95.3} & - & -    \\
\rowcolor{cyan!10} Dublin\textsubscript{fixed\_res} & \textbf{77.8} & \textbf{97.1} & 62.2  & 94.9  & \textbf{77.7/84.2}  & \textbf{347.3} \\
\rowcolor{cyan!10} Dublin\textsubscript{variable\_res} & \textbf{77.8} & \textbf{97.1}  & \textbf{65.7} & 94.9  & \textbf{77.7/84.2}  & \textbf{347.3}  \\
\midrule
SOTA with Spl. Pipelines  & \begin{tabular}[c]{@{}c@{}}(LyLMv3)\\ 92.08\end{tabular} & \begin{tabular}[c]{@{}c@{}}(UDOP)\\ 97.6\end{tabular} & \begin{tabular}[c]{@{}c@{}}(UDOP)\\ 85.5\end{tabular} & \begin{tabular}[c]{@{}c@{}}(UDOP)\\ 96.00\end{tabular} & \begin{tabular}[c]{@{}c@{}}(TIE)\\ 81.6/86.2\end{tabular} & \begin{tabular}[c]{@{}c@{}}(LyT5-large)\\ 344.1\end{tabular} \\
\cmidrule(lr){2-7}
BERT\textsubscript{large}/T5 (Text Baseline)& 65.63 & 90.25  & 74.4  & 89.92  & -  & -    \\
\bottomrule
\end{tabular}
\caption{
Performance on IE, doc classification, WebSRC and VisualMRC. Higher the better. LyLMv3: LayoutLMv3 \cite{huang2022layoutlmv3}, LyT5-large: LayoutT5-large \cite{kembhavi2016diagram}.
}
\label{tab:mixdatasets}
\end{table*}

Among the pixel-based models, we achieve state-of-the-art (SOTA) performance on AI2D, OCR-VQA, RefExp, DocVQA, InfoVQA, and CORD datasets. Notably, we stand as the global SOTA on AI2D, OCR-VQA, and RefExp, surpassing even current SOTA models that rely on specialized pipelines. Our performance on Widget Captioning, Screen2Words, TextCaps, and RVL-CDIP tasks remains highly competitive with the SOTA pixel-based models. However, we acknowledge that there is room for improvement in ChartQA performance. This could potentially be achieved by incorporating charts and diagrams into the pretraining data. 

\begin{table}[t!]
\centering
\small
\begin{tabular}{lccc}
\toprule
\multirow{2}{*}{Model} & \multicolumn{3}{c}{Table QA/NLI} \\
\cmidrule(lr){2-4}  
& \textbf{WTQ} & \textbf{TabFact}  & \textbf{WikiSQL}      \\
\midrule
Metrics    & EM           & Accuracy    & EM             \\
\midrule
\rowcolor{cyan!10} Dublin\textsubscript{fixed\_res}     & 25.7 & \textbf{73.54}   & 75.3   \\
\rowcolor{cyan!10} Dublin\textsubscript{variable\_res}     & \textbf{29.7} & 72.9 & 75.3     \\
\midrule
\begin{tabular}[c]{@{}c@{}}(SOTA w/)\\ Spl. pipelines\end{tabular}                                   & \begin{tabular}[c]{@{}c@{}}(UDOP)\\ 47.2\end{tabular} & \begin{tabular}[c]{@{}c@{}}(UDOP)\\ 78.9\end{tabular} & \begin{tabular}[c]{@{}c@{}}(TAPEX)\\ \textbf{89.2} \end{tabular}  \\
\cmidrule(lr){2-4}
\begin{tabular}[c]{@{}c@{}}(BART)\\ Text Baseline\end{tabular}& 38.0       & 76.0 & 85.8   \\


\bottomrule
\end{tabular}
\caption{
Performance on Table QA and NLI. Higher the better.
}
\label{tab:TableQA}
\end{table}

For datasets such as FUNSD, Deepform, WebSRC, VisualMRC, WTQ, and TabFact, WikiSQL and Squad1.1 pixel-based baselines were not previously established. We are the first to explore the potential of pixel-based models on these tasks. Notably, on VisualMRC, an abstractive QA task on document images, our model achieves global SOTA performance.  In Squad1.1, we create a pixel-based baseline achieving 77.7/84.2 as EM/F1 score whereas BART \cite{lewis2019bart} is at 86.44/93.04 and specialized pipeline (ANNA \cite{jun2022anna}) is 90.6/96.7. While our model may currently lag behind the specialized pipelines in FUNSD, DeepForm, WebSRC, WTQ, TabFact, WikiSQL and Squad1.1, this disparity can be attributed to the specialized pipelines' use of different modalities. For example, the TIE model \cite{zhao2022tie}, which is the global SOTA for WebSRC, leverages a specialized pipeline explicitly designed for WebSRC by combining Graph Attention Network and Pretrained Language Model to exploit topological and spatial structures. LayoutLMv3 \cite{huang2022layoutlmv3} and UDOP \cite{udop} models rely on OCR for their superior performance and the TAPEX model uses special architecture for table QA \cite{liu2022tapex}. Nonetheless, our pixel-based model shows promising potential in these tasks, and further exploration may yield improvements in performance.

\section{Conclusion}
\label{sec:conclusion}
We have presented DUBLIN,  a transformer-based encoder-decoder model for visual document understanding that can analyze both text and visual elements in document images. Evaluation on diverse downstream tasks show that it achieves competitive or superior performance compared to the existing state-of-the-art models. 

Our work shows that DUBLIN is a versatile and robust model that does not rely on external OCR systems and can be finetuned in an end-to-end fashion. We also introduce a new evaluation setup on text-based datasets by rendering them as images. While this is an unfair comparison as text-based models are expected to perform better for these tasks, this also serves as a challenging baseline for benchmarking VDU models. 

\section{Limitations}
Despite the promising results of our work, we recognize some limitations that we intend to overcome in future research. Our model has limited testing and evaluation on multilingual datasets. This may affect its applicability across languages and domains. Another limitation is the absence of evaluation for potential biases and other responsible AI issues that may emerge from the data or the text generation process. Additionally, we face the challenge of not being able to release the data and the model because of privacy reasons. Finally, our experiments were costly and required a total compute of 86000 GPU hours (which includes all failed experiments as well), which has an environmental impact as well. We aspire to find more efficient and sustainable ways to train and evaluate our model in the future.

\bibliography{custom}
\bibliographystyle{acl_natbib}
\appendix
\section*{APPENDIX}
\section{Pretraining Data}
\label{A:pretrainingData}
\textbf{CCNews 200M}~~~We use this dataset to obtain document images, texts, and bounding box coordinates in various web domains and languages. This is done by scraping the URLs from the CCNews 200M dataset \cite{commoncrawl} using the method outlined in CCNet \cite{wenzek-etal-2020-ccnet}  followed by rendering the HTML pages as screenshots and storing the document texts and their corresponding bounding boxes with the help of the Selenium library. We use samples from this dataset in all our pretraining tasks.

\noindent \textbf{Google NQ Dataset}~~~This is a publicly available dataset \cite{googleNQ} based on open domain question answering. It contains around 307k training samples, along with the URL/webpage link for each sample. We scrape the webpage content using the HTML URLs. The webpage content is rendered as an image with the question added at the top. The question will also be used as a prefix for the decoder. We train our model on this dataset on the Rendered Question Answering task. 

\noindent \textbf{Microsoft Bing QA Dataset}~~~We leverage a proprietary Bing QA dataset to obtain question-answer pairs along with their passage in English. We randomly sample question-answer pairs from search engine and render their passages and questions in a similar way as we did for the Google NQ dataset. In order to make our model's generalization ability better over different kinds of texts, we render the text with random font size, color, and style using the Google Fonts library. We use this dataset for the Rendered QA task

\noindent \textbf{Synthetic Table Structure QA Dataset}~~~In order to teach the model how to understand the table structure, we curate Synthetic Table QA dataset by randomly selecting 1 million webpages that contain tables and using Selenium to extract the HTML table elements from these webpages. To further enhance our training dataset, we perform data augmentation by employing five different CSS styles for rendering the HTML representation of each table as an image. These styles encompass various attributes such as border, font size, table separators, background, and text color. We devise this task of training the model to recognize table structure in the document images. During the training process, for each table, we randomly select one out of the five available styles. This ensured a diverse range of table appearances for our model to learn from. To generate synthetic questions and answers, we developed eleven distinct templates. These templates, reminiscent of SQL-like queries, were designed to reflect the content and format of the tables. An example template is as follows: "What is the value in the cell in the \texttt{[column\_name]} column, where the row contains \texttt{[row\_content]}?" Further elaboration on the templates and additional details can be found in Appendix~\ref{A:synethicTableQA}. 

\section{Pretraining Task}
\label{A:pretraining_tasks}
\begin{figure}[h]
\includegraphics[width=0.45\textwidth]{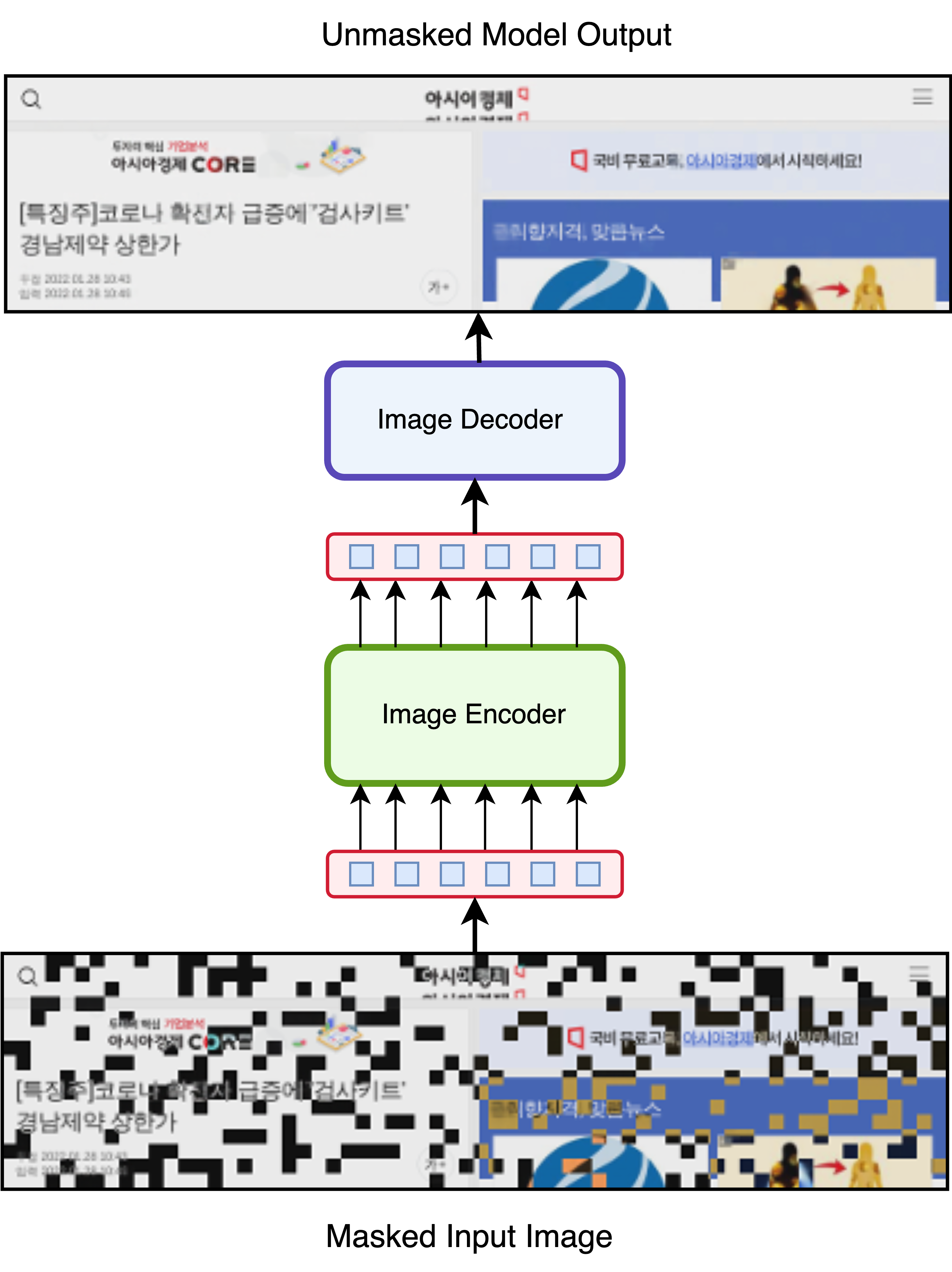}
\caption{Illustration of the MAE task with the masked image with model predictions inverted to better understand the masked patches.}
\label{fig:MAEtask}
\end{figure}

\textbf{Masked Autoencoding Task.}~~~Inspired by ViT-MAE, we use the MAE task. We mask out 15\% patch tokens of the image randomly in a similar fashion as was suggested in VIT-MAE \cite{he2021masked}. The task is to reconstruct the masked patches in the original image. We use 1-D fixed sinusoidal position embeddings to inject order information for the MAE task.
The image encoder and decoder are trained using a normalized mean squared error (MSE) pixel reconstruction loss, which measures the difference between the normalized target image patches and the reconstructed patches. This loss is specifically calculated for the masked patches. For a better understanding, Figure~\ref{fig:MAEtask} illustrates the MAE task, depicting the input image with masks and inverted predictions (inverted predictions are shown in the input image just for illustration and not added in the actual masked input image).

\section{Finetuning Datasets}
\label{A:finetuningdatasets}
\textbf{DocVQA}~~~DocVQA dataset \cite{mathew2021docvqa} focuses on question-answering tasks using single-page excerpts from real-world industry documents that include printed, handwritten and digital documents. The questions in this dataset often require understanding and processing various elements such as images, free text, tables, lists, forms, or a combination of these components. 

\noindent \textbf{InfographicsVQA}~~~
The InfographicVQA dataset \cite{mathew2021infographicvqa} contains questions that are specifically targeted at Infographics that can be found online. The inclusion of large images with extreme aspect ratios is one distinguishing feature of this dataset. Answering questions about visualized data found in a variety of Infographics is part of the task. The information needed to answer these questions can be presented using a variety of elements, including text, plots, graphs, or infographic layout components. 

\noindent \textbf{WebSRC}~~~WebSRC, also known as Web-based Structural Reading Comprehension, is a dataset consisting of 440,000 question-answer pairs \cite{websrc}. These pairs were collected from a diverse collection of 6,500 web pages. Each entry in the dataset includes not only the questions and answers but also the HTML source code, screenshots, and metadata associated with the respective web page. Answering questions in the WebSRC dataset requires a certain level of understanding of the structure of the web page. The answers can take the form of specific text excerpts, Key Information Extraction (KIE), or table question answering. To assess the performance on this dataset, we use metrics such as Exact Match (EM) and F1 score (F1). The training and development datasets are obtained using the official split provided by the authors. However, it's important to note that the authors have not released the testing set, so the results are solely based on the development set. 

\noindent \textbf{DeepForm}~~~We make use of the Key Information Extraction (KIE) dataset DeepForm \cite{Deepform}, which includes important election finance-related documents. The goal of this dataset is to extract crucial data from advertising disclosure forms submitted to the Federal Communications Commission (FCC), such as contract numbers, advertiser names, payment amounts, and air dates. Instead of the query, we provide the "Key" to the text decoder for the model to extract information from the image. 

\noindent \textbf{SQuAD1.1}~~~To evaluate our model's extractive question-answering performance, we fine-tune it on the SQuAD dataset \cite{rajpurkar2016squad}. We render this dataset as images on the fly, choosing a random font text, font style, etc., for each data point to maintain diversity and to test that, at inference time, the model is not biased toward answering questions from documents that all look a certain way but rather diverse in their fonts, styles, etc. The SQuAD dataset consists of over 100,000 question-answer pairs for over 500 articles. Given a question and its corresponding context paragraph, the task is to extract the span of text that contains the answer to the question. We follow the standard evaluation metrics for this dataset, including Exact Match (EM) and F1 score (F1), which measure the model's ability to output an answer that exactly matches the ground truth and its overlap with the ground truth, respectively.  By evaluating this widely used benchmark, we can compare the performance of our model against the state-of-the-art approaches in extractive question answering. 

\noindent \textbf{WikiTable}~~~WikiTableQuestions dataset \cite{pasupat2015compositional} utilized in this study focuses on question answering using semi-structured HTML tables obtained from Wikipedia. The authors specifically aimed to provide challenging questions that require multi-step reasoning on a series of entries within the given table, involving operations such as comparison and arithmetic calculations. We use the table images provided by the DUE Benchmark. 

\noindent \textbf{TabFact}~~~TabFact dataset includes entailed and refuted statements corresponding to a single row or cell to investigate fact verification using semi-structured evidence from clean and straightforward tables sourced from Wikipedia \cite{chen2020tabfact}. Despite the task's binary classification nature, it presents challenges that go beyond simple categorization. The task requires sophisticated linguistic and symbolic reasoning to achieve high accuracy. We pass the table image to the image encoder and expect a binary output from the text decoder for this table fact verification task. 

\noindent \textbf{WikiSQL}~~~WikiSQL is a large crowd-sourced dataset consisting of 80,654 meticulously annotated examples of questions and corresponding SQL queries \cite{zhong2017seq2sql}. These examples are derived from 24,241 tables extracted from Wikipedia.  This dataset mainly focuses on translating text to SQL. However, given our model's focus on answering questions based on documents, we transformed the denotations of this dataset into question-answer pairs in a natural language format. We rendered the tables as images by converting the table's JSON to HTML and then obtaining their screenshots in a similar fashion as described for the synthetic table structure QA dataset. 

\noindent \textbf{AI2D}~~~
AI2 Diagrams (AI2D) is a comprehensive dataset consisting of over 5000 science diagrams typically found in grade school textbooks, along with more than 150,000 annotations, including ground truth syntactic parses and over 15,000 corresponding multiple choice questions \cite{kembhavi2016diagram}. The diagrams cover a wide range of scientific topics, such as geological processes, biological structures, and more. The multiple-choice questions are based on the science diagrams and are designed to test students' comprehension of the content. The dataset provides only train and test splits, with 1 percent of the train split set aside for validation. 

\noindent \textbf{FUNSD}~~~FUNSD is a dataset in English for understanding forms in noisy scanned documents \cite{jaume2019funsd}. The FUNSD dataset contains 199 real, scanned forms with full annotations, comprising 9,707 labeled semantic entities across 31,485 words. The dataset is split into 149 samples for training and 50 samples for testing. The task involves semantic entity recognition, where each word is labeled with a category: question, answer, header, or other, using BIO tagging. To handle recurring entity names within a document, bounding boxes are drawn around the entities in the query image. The model is prompted with the question "Semantic label for this entity: <entity\_name> A) b-header B) i-header C) b-question D) i-question E) b-answer F) i-answer G) other" to make predictions. The evaluation metric is the entity-level F1 score. 

\noindent \textbf{CORD}~~~ CORD~\cite{Park2019CORDAC} is an English receipt dataset designed for key information extraction. It consists of 800 receipts for training, 100 for validation, and 100 for testing, with each receipt containing a photo and OCR annotations. The dataset defines 30 fields across 4 categories, and the task is to label each word with the appropriate field.  Official OCR annotations are utilized in the dataset. To handle recurring entity names within a document, bounding boxes are drawn around entities in the query image. The model is prompted with the question "What is the category for this entity: <entity\_name>" for making predictions. The evaluation metric used is the entity-level F1 score. 

\noindent \textbf{RefExp}~~~Referring expression component retrieval data (RefExp) is a dataset for the task of retrieving the UI component that a natural language expression refers to from a set of UI components detected on the screen \cite{bai2021uibert}. For example, given a UI image and an expression such as “Red button on the top”, the goal is to identify the UI component that matches the expression. Each sample in RefExp contains a UI image and a referring expression of a UI element on it.

\noindent \textbf{Widget Captioning}~~~The task of image captioning for widgets is addressed by the Widget Captioning dataset~\cite{li2020widget}. The dataset consists of app screenshots with a single widget (e.g. a button or a scroll bar) marked by a bounding box. The goal is to generate a caption that explains the functionality of the widget (e.g. find location). The dataset was generated by human workers and has 162,859 language phrases for 61,285 UI elements from 21,750 different UI screens.

\noindent \textbf{Screen2Words}~Screen2words dataset is a collection of app screenshots and their language summaries \cite{wang2021screen2words}. It is a large-scale dataset with more than 112k summaries for 
~22k different UI screens. The summaries were created by human workers and they explain the functionality of the page. The task is to generate a summary for an app screenshot that captures the page's functionality.

\noindent \textbf{ChartQA}~~~ChartQA is a large scale benchmark VQA dataset with 9.6K questions based on charts written by humans with 23.1K questions created from human-written chart summaries based on charts, i.e. visual representations of tabular data \cite{masry2022chartqa}. 

\noindent \textbf{OCR-VQA}~~~OCR-VQA~\cite{ocr-vqa} is a dataset for visual question answering by reading text in images. It contains images of book covers and questions based on book metadata such as title, author, genre, etc. The dataset comprises of 207,572 book cover images and more than 1 million question-answer pairs about these images.

\noindent \textbf{TextCaps}~~~We use TextCaps, a natural image captioning dataset, to study how to understand text in the context of an image. TextCaps contains 145k captions for 28k images. This dataset challenges a model to recognize text, relate it to its visual context, and decide what part of the text to copy or paraphrase, which requires spatial, semantic, and visual reasoning between multiple text tokens and visual entities, such as objects \cite{sidorov2020textcaps}. 

\noindent \textbf{RVL-CDIP}~~~The RVL-CDIP dataset, a benchmark document classification dataset~\cite{harley2015evaluation}, comprises 400,000 gray-scale images of English documents. The images are divided into 16 classes, with each class containing 25,000 images. The dataset poses a single-label multi-class classification task, where the model is prompted with the question "Classify the given document image" to predict the appropriate class among the 16 document categories. The evaluation metric for this task is the overall classification accuracy.

\noindent \textbf{VisualMRC}~~~The Visual MRC dataset is designed to facilitate the task of abstractive Question Answering (QA) in the context of document images \cite{tanaka2021visualmrc}. The primary objective of this dataset is to challenge machine learning models to comprehend the content of a document image based on a given question and generate a coherent and accurate abstractive answer. The evaluation metric used is CIDEr score.

\begin{figure}
    \centering
    \fbox{\includegraphics[width=1\linewidth]{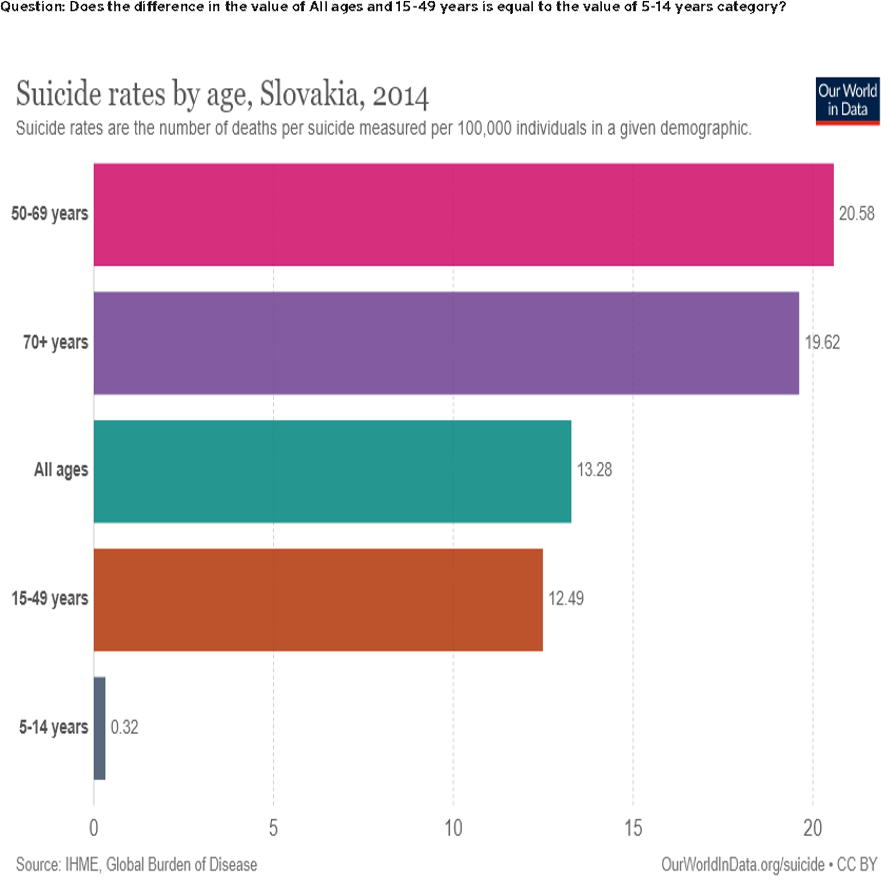}}
    \caption{Question rendered on top of the document image. }
    \label{fig:questionrendered}
\end{figure}

We append the question/key visually rendered onto the document image itself as can be seen in the Figure~\ref{fig:questionrendered}.

\section{Variable Resolution Scaling}
\begin{figure}
    \centering
    \includegraphics[width=1\linewidth]{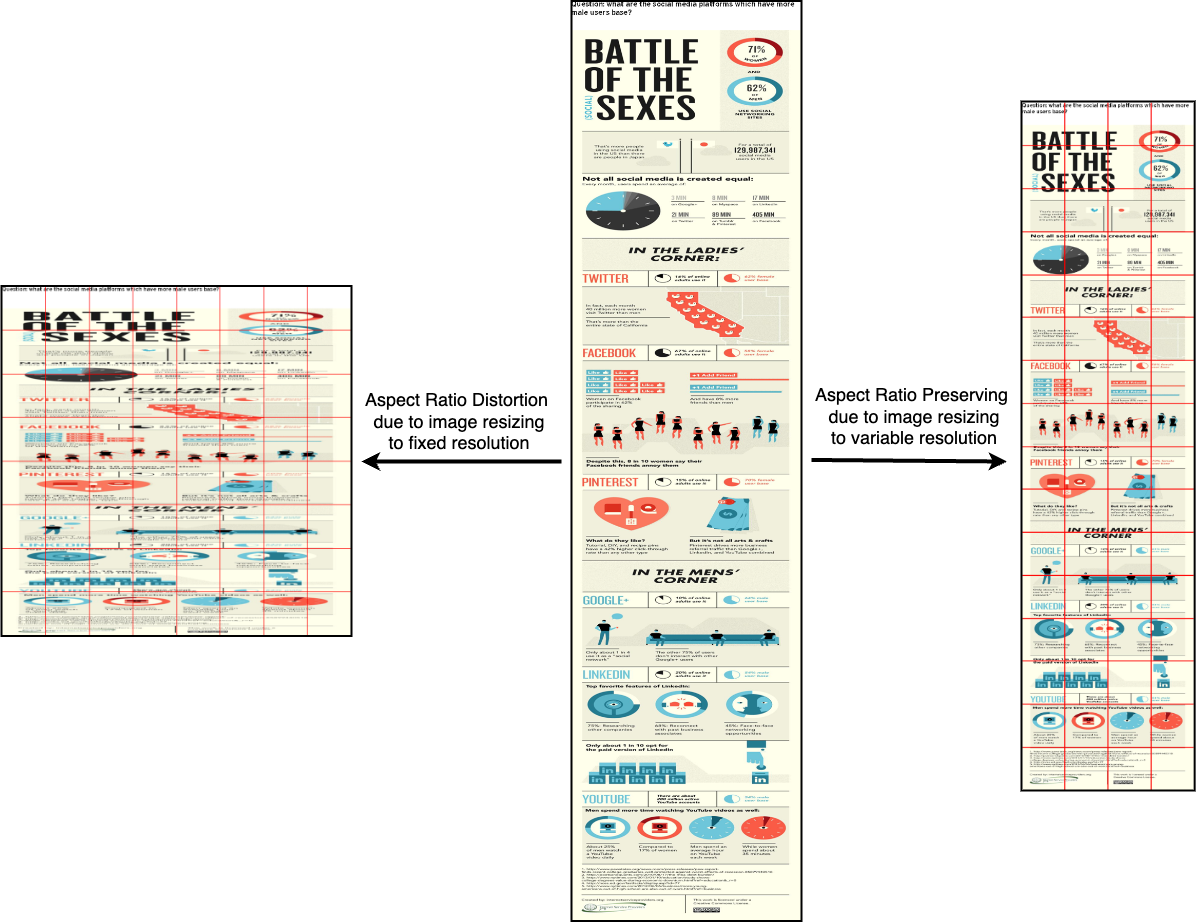}
    \caption{Illustration to show a comparison between variable resolution and typical fixed resolution approaches. Both inputs are pre-processed differently for a target of 64 patches. Suppose the original image is 1000$\times$200 (aspect ratio=5), we resize it to make the aspect ratio 4, the closest even power of 2. The image becomes 448$\times$112 for variable resolution but 224$\times$224 for fixed resolution.}
    \label{fig:varres}
\end{figure}

Figure \ref{fig:varres} compares our variable resolution and the typical fixed resolution methods. Our variable input resolution preserves the aspect ratio, while the fixed resolution input distorts the image and loses information along the longer side. Our variable resizing approach improves our models’ performance on datasets with longer documents, such as InfographivsVQA, DocVQA, and Deepform.

\section{Finetuning Hyperparameters}
\label{a:hyperparams}
For all finetuning experiments, we keep warmup steps constant at 1000 and weight decay at 0.01. Table~\ref{tab:hyperparams} contains the list of batch size and learning rate for finetuning on different datasets.
\begin{table}[h]
\centering
\begin{tabular}{lcc}
\toprule
\textbf{Datasets}                                 & \textbf{\begin{tabular}[c]{@{}c@{}}Batch\\ Size\end{tabular}} & \begin{tabular}[c]{@{}c@{}} \textbf{Learning} \\ \textbf{Rate} \end{tabular}\\
\midrule
\begin{tabular}[l]{@{}l@{}}OCR-VQA, WebSRC,\\ TextCaps, Squad, RefExp\end{tabular}  & 64                  & 1e-05                  \\
RVL-CDIP                                          & 256                 & 2e-05                  \\
All remaining datasets                            & 16                  & 1e-05              \\
\bottomrule
\end{tabular}
\caption{Hyperparameters for fine-tuning experiments.}
\label{tab:hyperparams}
\end{table}

\section{Model Results}

\begin{figure}
    \centering
    \fbox{\includegraphics[width=1\linewidth]{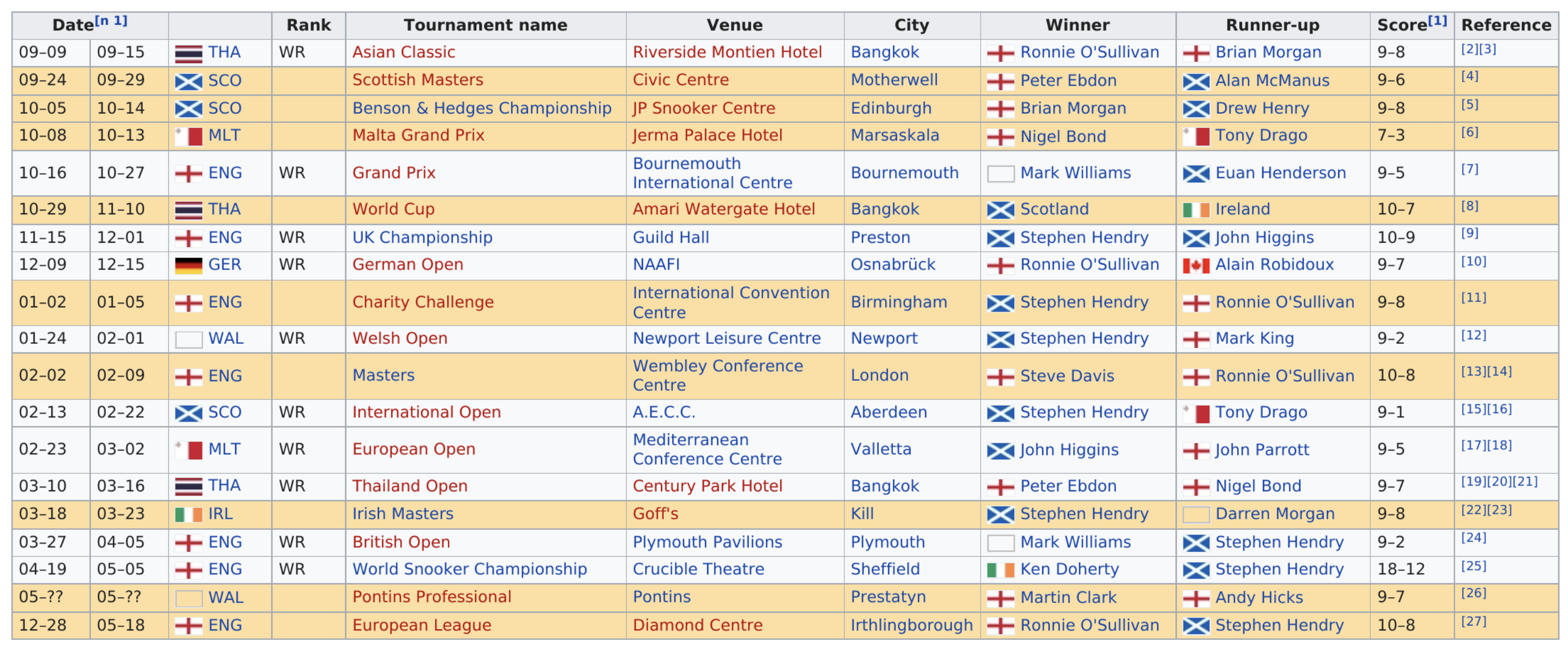}}
    \caption{Case 1 \\
    \texttt{Question:  What is the name of the first venue on this list?  \\
    DUBLIN's Answer: Riverside Montien Hotel \\
    Gold Answer: Riverside Montien Hotel}}
    \label{fig:case3wtq}
\end{figure}

\begin{figure}
    \centering
    \fbox{\includegraphics[width=\linewidth]{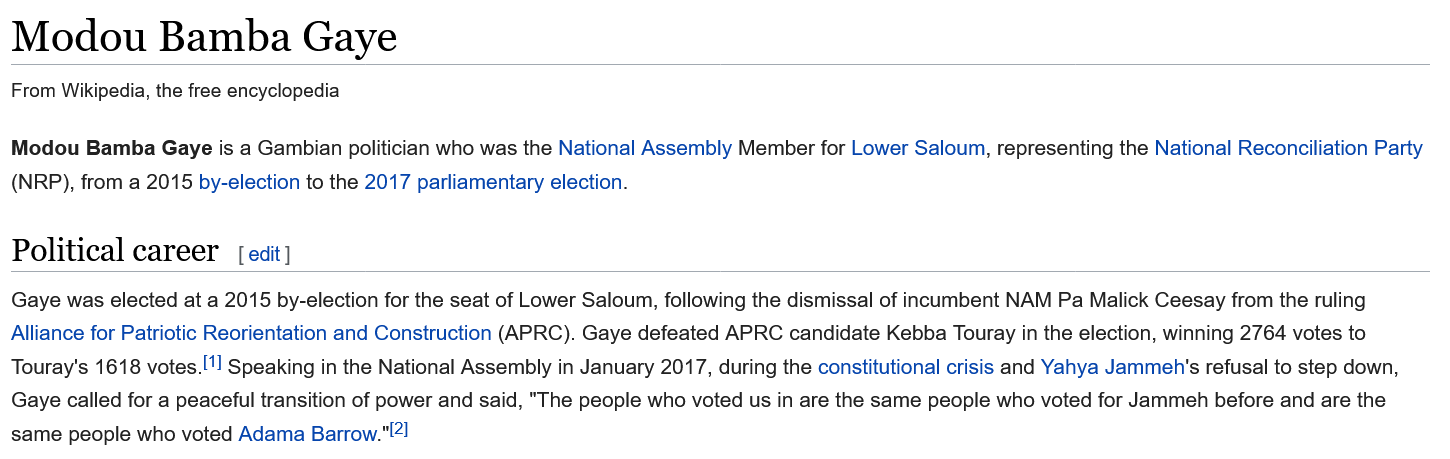}}
    \caption{Case 2 \\
    \texttt{Question:  When was Gaye elected for the seat of Lower Saloum?  \\
    DUBLIN's Answer: Gaye was elected at a 2015 by-election. \\
    Gold Answer: In 2015}}
    \label{fig:case2mrc}
\end{figure}

\begin{figure}
    \centering
    \fbox{\includegraphics[width=1\linewidth]{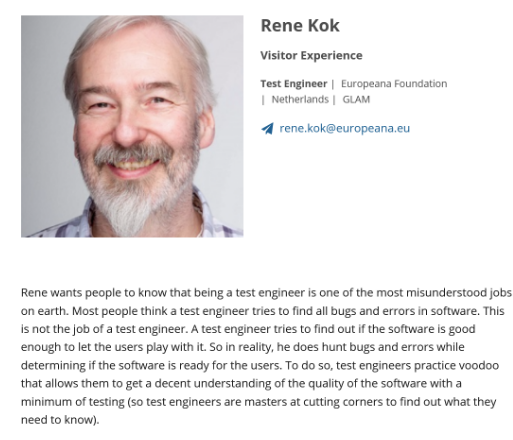}}
    \caption{Case 3 \\
    \texttt{Question:  What does Rene want people to know about being a test engineer? \\
    DUBLIN's Answer: He wants people to know that being a test engineer is one of the most misunderstood jobs on earth. \\
    Gold Answer: That being a test engineer is one of the most misunderstood jobs on earth.}}
    \label{fig:case1mrc}
\end{figure}

Figures \ref{fig:case3wtq}, \ref{fig:case2mrc}, and \ref{fig:case1mrc} show some examples of our model predictions compared to the gold answers for different images and questions.

\section{Synthetic Table Question Answering Dataset}
\label{A:synethicTableQA}
\begin{table*}[!hb]
\centering
\begin{tabular}{p{0.45\linewidth} | p{0.45\linewidth}}
\toprule
\multicolumn{1}{c|}{\textbf{Template}}                                                                & \multicolumn{1}{c}{\textbf{Example}}                                             \\
\midrule
What is the cell value in row \texttt{{[}row\_number{]}}  and column \texttt{{[}column\_number{]}}?                      & What is the cell value in row 3 and column 2?                                    \\
\rowcolor{gray!20} What is the cell value in column \texttt{{[}column\_number{]}} and row \texttt{{[}row\_number{]}}?                       & What is the cell value in column 7 and row 2?                                    \\
   What does the cell in the row \texttt{{[}row\_number{]}} and column \texttt{{[}column\_number{]}} contain?               & What does the cell in row 4 and column 9 contain?                                \\
\rowcolor{gray!20} What does the cell in column \texttt{{[}column\_number{]}} and row \texttt{{[}row\_number{]}} contain?                   & What does the cell in column 1 and row 3 contain?                                \\
What is the cell value in column \texttt{{[}column\_name{]}} and row \texttt{{[}row\_number{]}}?                         & What is the cell value in column "Price" and row 4?                              \\

\rowcolor{gray!20} What is the value of cell where column is \texttt{{[}column\_name{]}} and row number is \texttt{{[}row\_number{]}}?      & What is the value of cell where column is "Address" and row number is 9?           
\\
  What is the value in the cell in {[}column ordinal{]} column where the row contains {[}row entry{]}? & What is the value in the cell in second column where the row contains "Mangoes"? \\
\rowcolor{gray!20} What is the value for {[}column 1st entries{]}?                                                      & What is the value for "City"?                                                    \\
How many rows are there in this table?                                                               & -                                                                                \\
\rowcolor{gray!20} How many columns are there in this table?                                                            & -      \\              
  What is the caption of the table? & -\\
  \bottomrule                            
\end{tabular}
\caption{SQL-like query templates for generating QA pairs for the synthetic table-based question answering dataset.}
\label{t:templatesTableQA}
\end{table*}

\end{document}